%% file: main_neural_babytalk.tex
\documentclass[10pt,twocolumn,letterpaper]{article}

\usepackage{cvpr}
\usepackage{times}
\usepackage{epsfig}
\usepackage{graphicx}
\usepackage{amsmath}
\usepackage{amssymb}
\usepackage{bm}
\usepackage{booktabs}
\usepackage{multirow}
\usepackage{enumitem}
\usepackage{amsmath,bm}
\usepackage{bbm, dsfont}

\usepackage[pagebackref=true,breaklinks=true,letterpaper=true,colorlinks,bookmarks=false]{hyperref}

\cvprfinalcopy 

\def\cvprPaperID{205} 

\ifcvprfinal\fi
\begin{document}

\title{Neural Baby Talk}

\author{Jiasen Lu$^1$\thanks{Equal contribution} \quad Jianwei Yang$^1$\footnotemark[1] \quad Dhruv Batra$^{1,2}$ \quad Devi Parikh$^{1,2}$ \\
$^1$Georgia Institute of Technology \quad $^2$Facebook AI Research\\
{\tt\small \{jiasenlu, jw2yang, dbatra, parikh\}@gatech.edu}
}

\maketitle

\begin{abstract}
We introduce a novel framework for image captioning that can produce natural language explicitly grounded in entities that object detectors find in the image. 
Our approach reconciles classical slot filling approaches (that are generally better grounded in images) with modern neural captioning approaches (that are generally more natural sounding and accurate). 
Our approach first generates a sentence `template' with slot locations explicitly tied to specific image regions. These slots are then filled in by visual concepts identified in the regions by object detectors. The entire architecture (sentence template generation and slot filling with object detectors) is end-to-end differentiable. 
We verify the effectiveness of our proposed model on different image captioning tasks. On standard image captioning and novel object captioning, our model reaches state-of-the-art on both COCO and Flickr30k datasets. We also demonstrate that our model has unique advantages when the train and test distributions of scene compositions -- and hence language priors of associated captions -- are different. Code has been made available at: \href{https://github.com/jiasenlu/NeuralBabyTalk}{https://github.com/jiasenlu/NeuralBabyTalk}.
\end{abstract}

\input{intro}
\input{relatedWork}
\input{method}
\input{experiment}
\input{conclusion}
\input{supp}

{\small
\bibliographystyle{ieee}
\bibliography{egbib}
}

\end{document}

%% file: intro.tex
\vspace{-10pt}
\section{Introduction}
Image captioning is a challenging problem 
that lies at the intersection of computer vision and natural language processing. 
It involves generating a natural language sentence that accurately summarizes the contents of an image. 
Image captioning is also an important first step towards real-world applications with significant practical impact, ranging from aiding visually impaired users
to personal assistants to human-robot interaction \cite{antolVQA, visdial}. 

State-of-art image captioning models today tend to be monolithic neural models, essentially of the ``encoder-decoder'' paradigm. Images are encoded into a vector with a convolutional neural network (CNN), and captions are decoded from this vector using a Recurrent Neural Network (RNN), with the entire system trained end-to-end. While there are many recent extensions of this basic idea to include attention~\cite{Xu2015show, fang2015captions, you2016image, yang2016encode,lu2016knowing}, it is well-understood that models still lack visual grounding (\ie, do not associate named concepts to pixels in the image).
They often tend to `look' at different regions than humans would and tend to copy captions from training data \cite{vqahat}.

\begin{figure}[t]
 \centering 
 \includegraphics[width=1\linewidth, scale=1, trim={0in 0in 0.0in 0.0in}, clip]{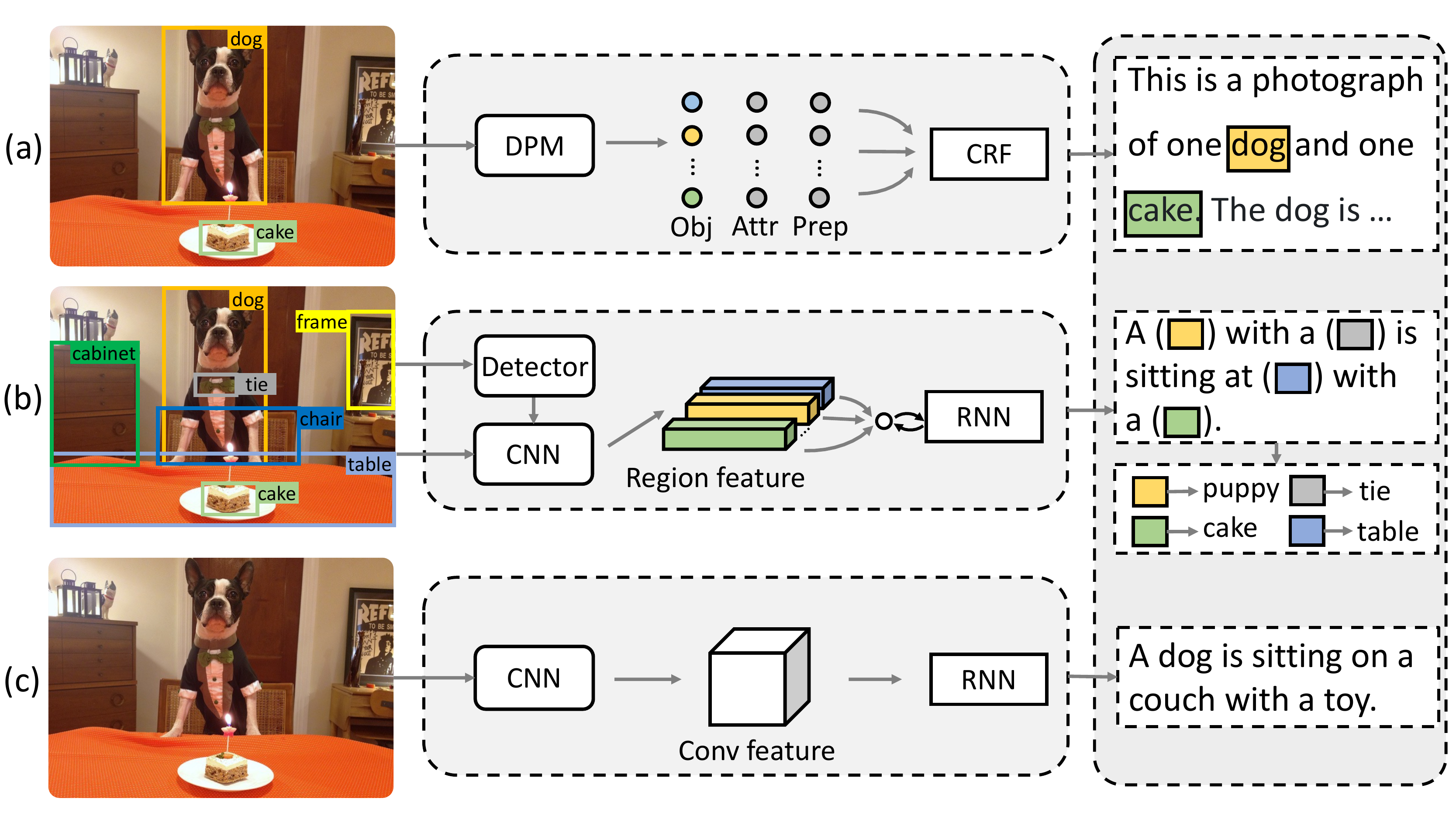}
 \caption{Example captions generated by (a) Baby Talk~\cite{kulkarni2013babytalk}, (c) neural image captioning~\cite{karpathy2015deep} and (b) our Neural Baby Talk approach. Our method generates the sentence ``template'' with slot locations (illustrated with filled boxes) explicitly tied to image regions (drawn in the image in corresponding colors). These slots are then filled by object detectors with concepts found in regions.  
}
\label{fig:teaser}
\vspace{-3mm}
\end{figure}

\begin{figure*}[t]
 \centering 
 \includegraphics[width=1\linewidth, scale=1, trim={0in 0in 0.0in 0.0in}, clip]{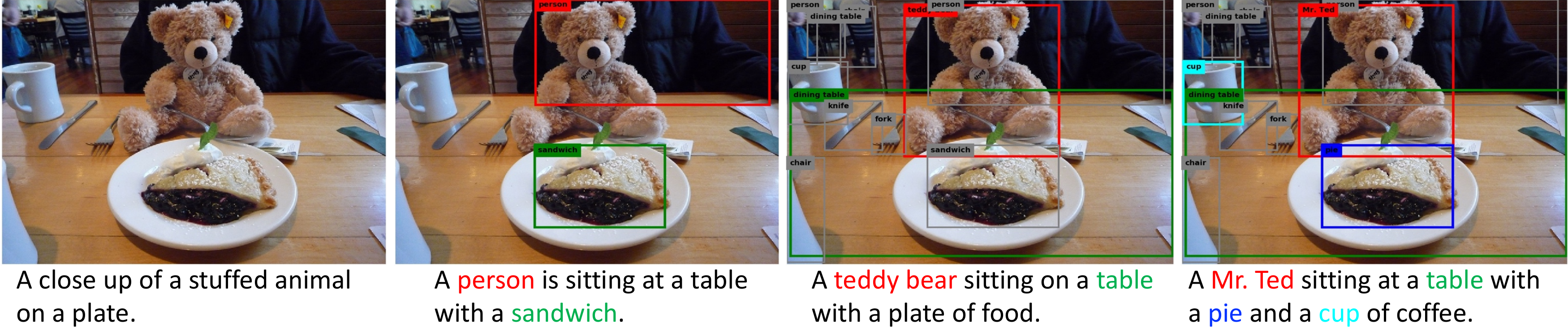}
 \caption{From left to right is the generated caption using the same captioning model but with different detectors: 1) No detector; 2) A weak detector that only detects ``person'' and ``sandwich''; 3) A detector trained on COCO~\cite{lin2014microsoft} categories (including ``teddy bear''). 4) A detector that can detect novel concepts (e.g. ``Mr. Ted'' and ``pie'' that never occurred in the captioning training data). Different colors show a correspondence between the visual word and grounding regions.}
\label{fig:demo1}
\vspace{-3mm}
\end{figure*}

For instance, in Fig.~\ref{fig:teaser} a neural image captioning approach~\cite{karpathy2015deep} describes the image
 as ``A dog is sitting on a couch with a toy." This is not quite accurate. But if one were to \emph{really} squint at the image, it (arguably) does perhaps look like a scene where a dog \emph{could} be sitting on a couch with a toy. 
It certainly is common to find dogs sitting on couches with toys. 
A-priori, the description is reasonable. 
That's exactly what today's neural captioning models tend to do -- produce generic \emph{plausible} captions based
on the language model\footnote{frequently, directly reproduced from a caption in the training data. } that match
a first-glance gist of the scene. While this may suffice for common scenes, images that differ 
from canonical scenes -- given the diversity in our visual world, there are \emph{plenty} of
such images -- tend to be underserved by these models.

If we take a step back -- do we really need the language model to do the heavy lifting in image captioning?
Given the unprecedented progress we are seeing in object recognition\footnote{e.g., 11\%
absolute increase in average precision in object detection in the COCO challenge in the last year.} 
(e.g., object detection, semantic segmentation, instance segmentation, pose estimation), 
it seems like the vision pipeline can certainly do better than rely on just a first-glance gist of the scene. 
In fact, today's state-of-the-art object detectors can successfully detect the table and cake in the image 
in Fig.~\ref{fig:teaser}(c)! The caption ought to be able to talk about the table and cake \emph{actually detected} as opposed to letting the language model hallucinate a couch and a toy simply because that sounds plausible.

Interestingly, some of the first attempts at image captioning~\cite{farhadi2010every, kulkarni2013babytalk, kuznetsova2012collective,mitchell2012midge} -- before the deep learning ``revolution'' -- 
relied heavily on outputs of object detectors and attribute classifiers to describe images. 
For instance, consider the output of Baby Talk~\cite{kulkarni2013babytalk} in Fig.~\ref{fig:teaser}, that used a slot filling 
approach to talk about all the objects and attributes found in the scene via a templated caption. The language 
is unnatural but the caption is very much grounded in what the model 
sees in the image. Today's approaches fall at the other extreme on the spectrum -- 
the language generated by modern neural image captioning approaches is much more natural but tends to be much less grounded in the image.

In this paper, we introduce Neural Baby Talk that reconciles these methodologies. It produces natural language \emph{explicitly} grounded in entities found by object detectors. It is a neural approach that generates a sentence 
``template'' with slot locations explicitly tied to image regions. These slots are then filled by object recognizers with concepts found in the regions. 
The entire approach is trained end-to-end.  This results in natural 
sounding and grounded captions. 

Our main technical contribution is a novel neural decoder for grounded image captioning. Specifically, at each time step, the model decides whether to generate a word from the textual vocabulary 
or generate a ``visual'' word. The visual word is essentially a token that will hold the slot for 
a word that is to describe a specific region in the image.
For instance, for the image
in Fig.~\ref{fig:teaser}, the generated sequence may be ``A $<$region$-$17$>$ is sitting at a $<$region$-$123$>$ with a $<$region$-3$$>$." 
The visual
words ($<$region$-$[.]$>$'s) are then filled in during a second stage 
that classifies each of the indicated regions (e.g., $<$region$-$17$>$$\rightarrow$puppy, 
$<$region$-$123$>$$\rightarrow$table), resulting in 
a final description of ``A puppy is sitting at a table with a cake.'' -- a free-form natural language description that is grounded in the image. One nice feature of our model is that it allows for different object detectors to be plugged in easily. As a result, a variety of captions can be produced for the same image using different detection backends. See Fig.~\ref{fig:demo1} for an illustration.

\noindent \textbf{Contributions:} Our contributions are as follows: 
\begin{itemize}[nolistsep]
\item We present Neural Baby Talk -- a novel framework for visually grounded image captioning that explicitly localizes objects in the image while generating free-form natural language descriptions. 
\item Ours is a two-stage approach that first generates a hybrid template that contains a mix of (text) words and slots explicitly associated with image regions, and then fills in the slots with (text) words by recognizing the content in the corresponding image regions.
\item We propose a robust image captioning task to benchmark compositionality of image captioning algorithms where at test time the model encounters images containing known objects but in novel combinations (e.g., the model has seen dogs on couches and people at tables during training, but at test time encounters a dog at a table). Generalizing to such novel compositions is one way to demonstrate image grounding as opposed to simply leveraging correlations from training data.
\item Our proposed method achieves state-of-the-art performance on COCO and Flickr30k datasets on the standard image captioning task, and significantly outperforms existing approaches on the robust image captioning and novel object captioning tasks.
\end{itemize}

%% file: relatedWork.tex
\section{Related Work}

Some of the earlier approaches generated templated image captions via slot-filling. 
For instance, 
Kulkarni \etal~\cite{kulkarni2013babytalk} detect objects, attributes, and prepositions, jointly reason about these through a CRF, and finally fill appropriate slots in a template. Farhadi \etal \cite{farhadi2010every} compute a triplet for a scene, and use this templated ``meaning'' representation to retrieve a caption from a database. \cite{kuznetsova2012collective,mitchell2012midge} use more powerful language templates such as a syntactically well-formed tree. These approaches tend to either produce captions that are relevant to the image but not natural sounding, or captions that are natural (\eg retrieved from a database of captions) but may not be sufficiently grounded in the image. 

Neural models for image captioning have been receiving increased attention in the last few years~\cite{kiros2014ICML, mao2014deep, chen2015mind, vinyals2015show, donahue2015long, karpathy2015deep}. State-of-the-art neural approaches include attention mechanisms~\cite{Xu2015show, fang2015captions, you2016image, yang2016encode, lu2016knowing, rennie2016self, Anderson2017up-down} that identify regions in the image to ``ground'' emitted words. In practice, these attention regions tend to be quite blurry, and rarely correspond to semantically meaningful individual entities (e.g., objects instances) in the image. Our approach grounds words in object detections, which by design identify concrete semantic entities (object instances) in the image. 

There has been some recent interest in grounding natural language in images. Dense Captioning \cite{densecap} generates descriptions for specific image regions. In contrast, our model produces captions for the entire image, with words grounded in concrete entities in the image. 
Another related line of work is on resolving referring expressions~\cite{kazemzadeh2014referitgame}
(or description-based object retrieval~\cite{plummer2015flickr30k,hu2016modeling,hu2016natural,rohrbach2016grounding} -- given a description of an object in the image, identify which object is being referred to) or referring expression generation~\cite{kazemzadeh2014referitgame,luo2017comprehension, mao2016generation, yu2016modeling} (given an object in the image, generate a discriminative description of the object). While the interest in grounded language is in common, our task is different. 

One natural strength of our model is its ability to incorporate different object detectors, including the ability to generate captions with novel objects (never seen before in training captions). In that context, our work is related to prior works on novel object captioning \cite{anne2016deep,venugopalan2016captioning,yao2017incorporating, anderson2016guided}. As we describe in Sec.~\ref{sec:noc_exp}, our method outperforms these approaches by 14.6\% on the averaged F1 score. 


%% file: method.tex
\section{Method}



Given an image $\bm{I}$, the goal of our method is to generate visually grounded descriptions $\bm{y}=\{y_1,\ldots,y_T\}$. Let $\bm{r}_{\bm{I}} = \{r_1, ..., r_N\}$ be the set of $N$ images regions extracted from $\bm{I}$. 
When generating an entity word in the caption, we want to ground it in a specific image region $r \in \bm{r}_{\bm{I}}$. Following the standard supervised learning paradigm, we learn parameters $\bm{\theta}$ of our model by maximizing the likelihood of the correct caption:
\vspace{-5pt}
\begin{equation}
\bm{\theta}^* = \arg \max_{\bm{\theta}} \sum_{(\bm{I},\bm{y})} \log p(\bm{y}\vert \bm{I};\bm{\theta})
\label{eq:1}
\vspace{-5pt}
\end{equation}

Using chain rule, the joint probability distribution can be decomposed over a sequence of tokens:

\vspace{-5pt}
\begin{equation}
p(\bm{y}|\bm{I}) = \prod_{t=1}^T p(y_t \vert \bm{y}_{1:t-1}, \bm{I})
\label{eq:2}
\vspace{-5pt}
\end{equation}

where we drop the dependency on model parameters to avoid notational clutter. We introduce a latent variable $r_t$ to denote a specific image region so that $y_t$ can explicitly ground in it. Thus the probability of $y_t$ is decomposed to:
\vspace{-5pt}
\begin{equation}
p(y_t \vert \bm{y}_{1:t-1}, \bm{I}) = p(y_t \vert r_t, \bm{y}_{1:t-1}, \bm{I}) p(r_t \vert \bm{y}_{1:t-1}, \bm{I})
\label{eq:3}
\vspace{-5pt}
\end{equation}

In our framework, $y_t$ can be of one of two types: a visual word or a textual word, denoted as $y^{vis}$ and $y^{txt}$ respectively. A visual word $y^{vis}$ is a type of word that is grounded in a specific image region drawn from $\bm{r}_{\bm{I}}$.  A textual word $y^{txt}$ is a word from the remainder of the caption. It is drawn from the language model , which is associated with a ``default'' sentinel ``region'' $\tilde{r}$ obtained from the language model~\cite{lu2016knowing} (discussed in Sec.~\ref{sec:visualword}). For example, as illustrated in Fig.~\ref{fig:teaser}, ``puppy'' and ``cake'' grounded in the bounding box of category ``dog'' and ``cake'' respectively, are visual words. While ``with'' and ``sitting'' are not associated with any image regions and thus are textual words.

\begin{figure}[t]
 \centering
 \includegraphics[width=1.0\linewidth, scale=1, trim={0in 0.0in 0in 0.0in}, clip]{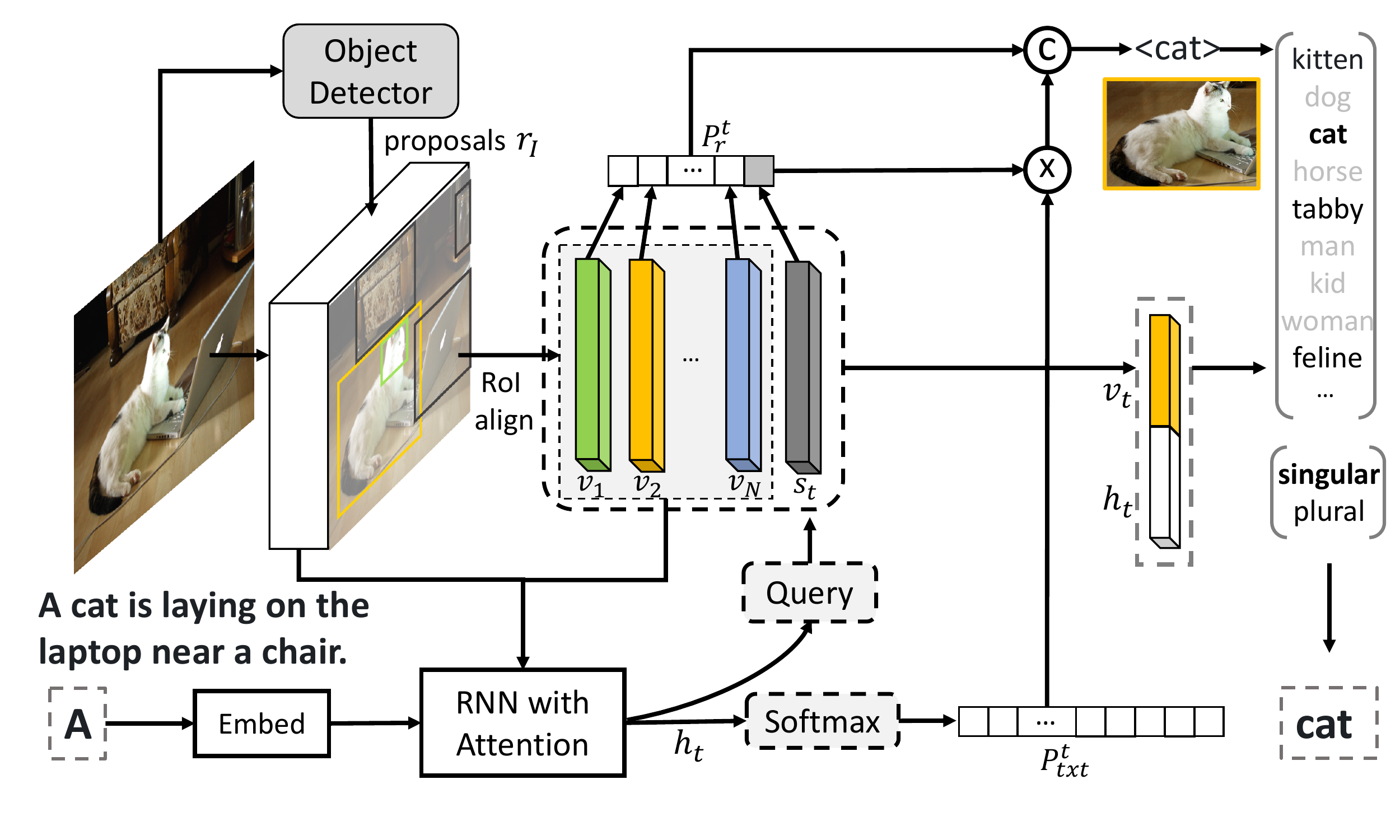} 
 \vspace{-20pt}
 \caption{One block of the proposed approach. Given an image, proposals from any object detector and current word ``A'', the figure shows the process to predict the next visual word ``cat''.}
\label{fig:overview}
\vspace{-3mm}
\end{figure}

With this, Eq.~\ref{eq:1} can be decomposed into two cascaded objectives. First, maximizing the probability of generating the sentence ``template''. 
A sequence of grounding regions associated with the visual words interspersed with the textual words can be viewed as a sentence ``template'', where the grounding regions are slots to be filled in with visual words.\footnote{Our approach is not limited to any pre-specified bank of templates. Rather, our approach automatically generates a template (with placeholders -- slots -- for visually grounded words), which may be any one of the exponentially many possible templates.} An example template (Fig.~\ref{fig:overview}) is ``A $<$region$-$2$>$ is laying on the $<$region$-$4$>$ near a $<$region$-$7$>$. Second, maximizing the probability of visual words $y_t^{vis}$ conditioned on the grounding regions and object detection information, e.g., categories recognized by detector. In the template example above, the model will fill the slots with `cat', `laptop' and `chair' respectively.

In the following, we first describe how we generate the slotted caption template (Sec.~\ref{sec:visualword}), and then how the slots are filled in to obtain the final image description (Sec.~\ref{sec:adp}). The overall objective function is described in Sec.~\ref{sec:obj} and the implementation details in Sec.~\ref{sec:detail}.

\subsection{``Slotted'' Caption Template Generation}
\label{sec:visualword}
Given an image $\bm{I}$, and the corresponding caption $\bm{y}$, the candidate grounding regions are obtained by using a pre-trained Faster-RCNN network \cite{ren2015faster}. To generate the caption ``template'', we use a recurrent neural network, which is commonly used as the decoder for image captioning \cite{mao2014deep,vinyals2015show}. At each time step, we compute the RNN hidden state $\bm{h}_t$ according to the previous hidden state $\bm{h}_{t-1}$ and the input $\bm{x}_t$ such that $\bm{h}_t = \mathrm{RNN}(\bm{x}_t, \bm{h}_{t-1})$. At training time, $x_t$ is the ground truth token (teacher forcing) and at test time is the sampled token $y_{t-1}$. Our decoder consists of an attention based LSTM layer \cite{rennie2016self} that takes convolution feature maps as input. Details can be found in Sec.~\ref{sec:detail}. To generate the ``slot'' for visual words, we use a pointer network \cite{vinyals2015pointer} that modulates a content-based attention mechanism over the grounding regions. Let $\bm{v_t} \in \mathcal{R}^{d \times 1}$ be the region feature of ${r}_t$, which is calculated based on Faster R-CNN. We compute the pointing vector with:
\begin{align}
u_i^t &= \bm{w}_{h}^T \tanh(\bm{W}_v \bm{v}_t + \bm{W}_{z} \bm{h}_t) 
\label{eq:4} \\
\bm{P}_{\bm{r}_{\bm{I}}}^t & = \textrm{softmax}(\bm{u}^t) 
\label{eq:5}
\end{align}
where $\bm{W}_v \in \mathbb{R}^{m \times d}$, $\bm{W}_z \in \mathbb{R}^{d \times d}$ and $\bm{w}_h \in \mathbb{R}^{d \times 1}$ are parameters to be learned. The $\textrm{softmax}$ normalizes the vector $\bm{u}^t$ to be a distribution over grounding regions $\bm{r}_{\bm{I}}$.

Since textual words $y_t^{txt}$ are not tied to specific regions in the image, inspired by~\cite{lu2016knowing}, we add a ``visual sentinel'' $\tilde{r}$ as a latent variable to serve as dummy grounding for the textual word. The visual sentinel can be thought of as a latent representation of what the decoder already knows about the image.
The probability of a textual word $y_t^{txt}$ then is:
\begin{equation}
p(y_t^{txt} \vert \bm{y}_{1:t-1}) = p(y_t^{txt} \vert \tilde{r}, \bm{y}_{1:t-1}) p(\tilde{r} \vert \bm{y}_{1:t-1})
\label{eq:6}
\end{equation}
where we drop the dependency on $\bm{I}$ to avoid clutter. 

We first describe how the visual sentinel is computed, and then how the textual words are determined based on the visual sentinel. Following~\cite{lu2016knowing}, when the decoder RNN is an LSTM \cite{hochreiter1997long}, the representation for visual sentinel $\bm{s}_t$ can be obtained by:

\vspace{-10pt}
\begin{eqnarray}
\bm{g}_t &=& \sigma \left( \bm{W}_{x} \bm{x}_t + \bm{W}_{h} \bm{h}_{t-1} \right)\\
\label{eq:7}
\bm{s}_t &=& \bm{g}_t \odot \tanh \left( \bm{c}_ t\right)
\label{eq:8}
\vspace{-10pt}
\end{eqnarray}
where $\bm{W}_x \in \mathbb{R}^{d\times d }$, $\bm{W}_h \in \mathbb{R}^{d\times d}$. $\bm{x}_t$ is the LSTM input at time step $t$, and $\bm{g}_t$ is the gate applied on the cell state $\bm{c}_t$. $\odot$ represents element-wise product, $\sigma$ the logistic sigmoid activation. Modifying Eq.~\ref{eq:5}, the probability over the grounding regions including the visual sentinel is: 
\vspace{-5pt}
\begin{equation}
\bm{P}_{\bm{r}}^t = \textrm{softmax}([\bm{u}^t; \bm{w}_{h}^T\tanh(\bm{W}_s \bm{s}_t + \bm{W}_{z} \bm{h}_t)])
\label{eq:9}
\vspace{-5pt}
\end{equation}
where $\bm{W}_s \in \mathbb{R}^{d \times d}$ and $\bm{W}_z \in \mathbb{R}^{d \times d}$ are the  parameters. Notably, $\bm{W}_z$ and $\bm{w}_{h}$ are the same parameters as in Eq.~\ref{eq:4}. $\bm{P}_{\bm{r}}^t$ is the probability distribution over grounding regions $\bm{r}_{\bm{I}}$ and visual sentinel $\tilde{r}$. The last element of the vector in Eq.~\ref{eq:9} captures $p(\tilde{r} \vert \bm{y}_{1:t-1})$.

We feed the hidden state $\bm{h}_t$ into a $\textrm{softmax}$ layer to obtain the probability over textual words conditioned on the image, all previous words, and the visual sentinel:
\vspace{-5pt}
\begin{equation}
\bm{P}_{txt}^t = \textrm{softmax}\left(\bm{W}_q \bm{h}_t \right)
\label{eq:10}
\vspace{-5pt}
\end{equation}
where $\bm{W}_q \in \mathbb{R}^{V \times d}$, $d$ is hidden state size, and $V$ is textual vocabulary size. 
Plugging in Eq.~\ref{eq:10} and $p(\tilde{r} \vert \bm{y}_{1:t-1})$ from the last element of the vector in Eq.~\ref{eq:9} into Eq.~\ref{eq:6} gives us the probability of generating a textual word in the template. %

\subsection{Caption Refinement: Filling in The Slots}
\label{sec:adp}
To fill the slots in the generated template with visual words grounded in image regions, we leverage the outputs of an object detection network. 
Given a grounding region, the category can be obtained through any detection framework \cite{ren2015faster}. But outputs of detection networks are typically singular coarse labels \eg ``dog''. Captions often refer to these entities in a fine-grained fashion \eg ``puppy'' or in the plural form ``dogs''. In order to accommodate for these linguistic variations, the visual word $y^{vis}$ in our model is a refinement of the category name by considering the following two factors:
First, determine the plurality -- whether it should be singular or plural. Second, determine the fine-grained class (if any). Using two single layer MLPs with ReLU activation $f(\cdot)$, we compute them with:

\vspace{-5pt}
\begin{align}
\bm{P}_{b}^t &= \textrm{softmax}\left(\bm{W}_b f_b \left( \left[\bm{v}_t; \bm{h}_t\right] \right) \right)\\
\bm{P}_{g}^t &= \textrm{softmax}\left(\bm{U}^T\bm{W}_g f_g \left(\left[\bm{v}_t; \bm{h}_t\right] \right) \right)
\vspace{-5pt}
\end{align}
$\bm{W}_b \in \mathbb{R}^{2 \times d}$,  $\bm{W}_g \in \mathbb{R}^{300 \times d}$ are the weight parameters. $\bm{U} \in \mathbb{R}^{300 \times k}$ is the glove vector embeddings \cite{pennington2014glove} for $k$ fine-grained words associated with the category name. The visual word $y_t^{vis}$ is then determined by plurality and fine-grained class (\eg, if plurality is plural, and the fine-grained class is ``puppy'', the visual word will be ``puppies'').
%

\subsection{Objective}
\label{sec:obj}
Most standard image captioning datasets (\eg COCO \cite{lin2014microsoft}) do not contain phrase grounding annotations, while some datasets do (\eg Flickr30k \cite{plummer2015flickr30k}).  Our training objective (presented next) can incorporate different kinds of supervision
-- be it strong annotations 
indicating which words in the caption are grounded in which 
boxes in the image, or weak supervision where objects are 
annotated in the image but are not aligned to words in the 
caption.
Given the target ground truth caption $\bm{y}_{1:T}^*$ and a image captioning model with parameters $\bm{\theta}$, we minimize the cross entropy loss:
\begin{equation}
\begin{aligned}
L(\bm{\theta})& =
 - \sum_{t=1}^T \log \Big(
\overbrace{p(y_t^* \vert \tilde{r}, \bm{y}_{1:t-1}^*) p(\tilde{r} \vert \bm{y}_{1:t-1}^*) \mathbbm{1}_{(y_t^* = y^{\textrm{txt}})}}^{\substack{\text{Textual word probability}}}
 + \\ 
&\underbrace{p\left(b_t^*, s_t^* \vert \bm{r}_t, \bm{y}_{1:t-1}^*\right)}_{\text{Caption refinement}}
\big( 
\underbrace{\dfrac{1}{m} \sum_{i=1}^m p\left(r_t^i \vert \bm{y}^*_{1:t-1}\right)\big) \mathbbm{1}_{(y_t^* = y^{\textrm{vis}})}}_{\substack{\text{Averaged target region probability}}} \Big)
\end{aligned}
\end{equation}
where $y_t^*$ is the word from the ground truth caption at time $t$. $\mathbbm{1}_{(y_t^* = y^{\textrm{txt}})}$ is the indicator function which 
equals to 1 if $y_t^*$ is textual word and 0 otherwise. $b_t^*$ and $s_t^*$ 
are the target ground truth plurality and find-grained class. 
$\{r_t^i\}_{i=1}^m \in \bm{r_I}$ are the target grounding regions of the visual word at time $t$. We maximize the averaged log probability of the target grounding regions. 

\textbf{Visual word extraction.} During training, visual words in a caption are dynamically identified by matching the base form of each word (using the Stanford lemmatization 
toolbox~\cite{manning-EtAl:2014:P14-5}) against a vocabulary 
of visual words (details of how to get visual word can be found in dataset Sec.~\ref{sec:experiment}). 
The grounding regions $\{r_t^i\}_{i=1}^m$ for a visual word $y_t$ is 
identified by computing the IoU of all boxes detected by the 
object detection network with the ground truth bounding box 
associated with the category corresponding to $y_t$. 
If the score exceeds a threshold of $0.5$ and the grounding region label
matches the visual word, the bounding boxes 
are selected as the grounding regions. E.g., given a target visual word ``cat'', if there are no proposals that match the target bounding box, the model predicts the textual word ``cat'' instead.

\subsection{Implementation Details}
\label{sec:detail}
\textbf{Detection model.} We use Faster R-CNN \cite{ren2015faster} with ResNet-101 \cite{he2015deep} to obtain region proposals for the image. We use an IoU threshold of 0.7 for region proposal suppression and 0.3 for class suppressions. A class detection confidence threshold of 0.5 is used to select regions. 

\textbf{Region feature.} We use a pre-trained ResNet-101 \cite{he2015deep} in our model. The image is first resized to $576 \times 576$ and we random crop $512 \times 512$ as the input to the CNN network. Given proposals from the pre-trained detection model, the feature $\bm{v}_i$ for region $i$ is a concatenation of 3 different features $\bm{v}_i = [\bm{v}_i^p; \bm{v}_i^l; \bm{v}_i^g]$ where $\bm{v}_i^p$ is the pooling feature of RoI align layer \cite{he2017mask} given the proposal coordinates,  $\bm{v}_i^l$ is the location feature and $\bm{v}_i^g$ is the glove vector embedding of the class label for region $i$. 
Let $x_\text{min}, y_\text{min}, x_\text{max}, y_\text{max}$ be the bounding box coordinates of the region $b$; $W_I$ and $H_I$ be the width and height of the image $I$. Then the location feature $\bm{v}_i^l$ can be obtained by projecting the normalized location $[\dfrac{x_\text{min}}{W_I}, \dfrac{y_\text{min}}{H_I}, \dfrac{x_\text{max}}{W_I}, \dfrac{y_\text{max}}{H_I}]$ into another embedding space.

\begin{figure}[t]
 \centering 
 \includegraphics[width=1.0\linewidth, scale=1, trim={0in 0.8in 0in 0.8in}, clip]{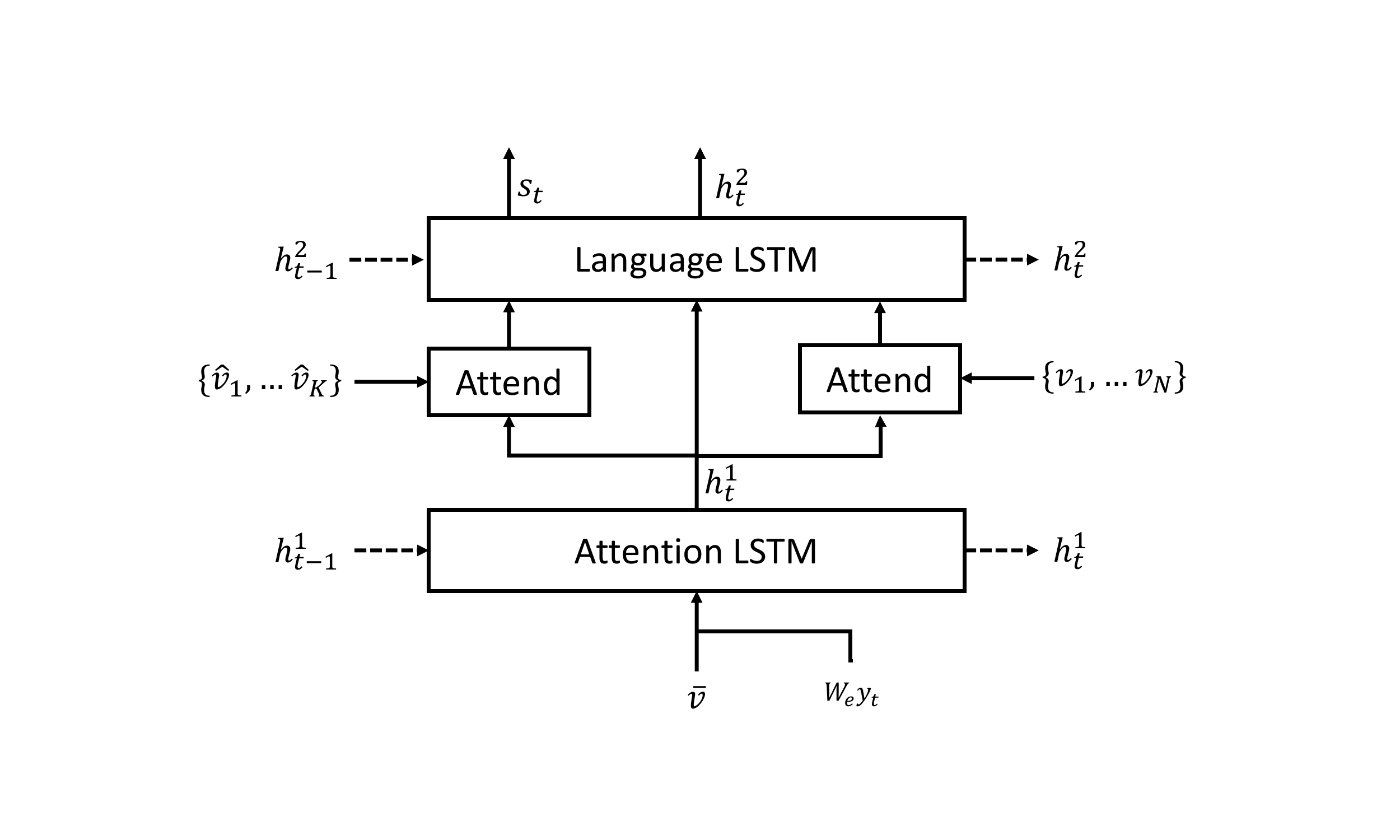}
\caption{Language model used in our approach.}
\label{fig:langugage_model}
\vspace{-3mm}
\end{figure}

\textbf{Language model.} We use an attention model with two LSTM layers~\cite{Anderson2017up-down} as our base attention model. 
Given $N$ region features from detection proposals $\bm{V} = \{\bm{v}_1, \ldots, \bm{v}_N\}$ and CNN features from the last convolution layer at $K$ grids $\bm{\hat{V}} = \{\bm{\hat{v}}_1, \ldots, \bm{\hat{v}}_K\}$, the language model has two separate attention layers shown in Fig~\ref{fig:langugage_model}. The attention distribution over the image features for detection proposals is:

\vspace{-5pt}
\begin{equation}
\begin{aligned}
\bm{z}_t &= \bm{w}_z^T \tanh \left(\bm{W}_v \bm{V} + (\bm{W_g\bm{h}_t}) \mathbbm{1}^T \right) \\
\bm{\alpha}_t &= \mathrm{softmax}(\bm{z}_t)
\end{aligned}
\vspace{-5pt}
\end{equation}
where $\bm{W}_v \in \mathbb{R}^{m \times d}$, $\bm{W}_g \in \mathbb{R}^{d \times d}$ and $\bm{w} \in \mathbb{R}^{d \times 1}$. $\mathbbm{1} \in \mathbb{R}^N$ is a vector with all elements set to 1. $\bm{\alpha}_t$ is the attention weight over $N$ image location features. 

\begin{figure*}[t]
 \centering 
 \includegraphics[width=1\linewidth, scale=1, trim={0in 0in 0.0in 0.0in}, clip]{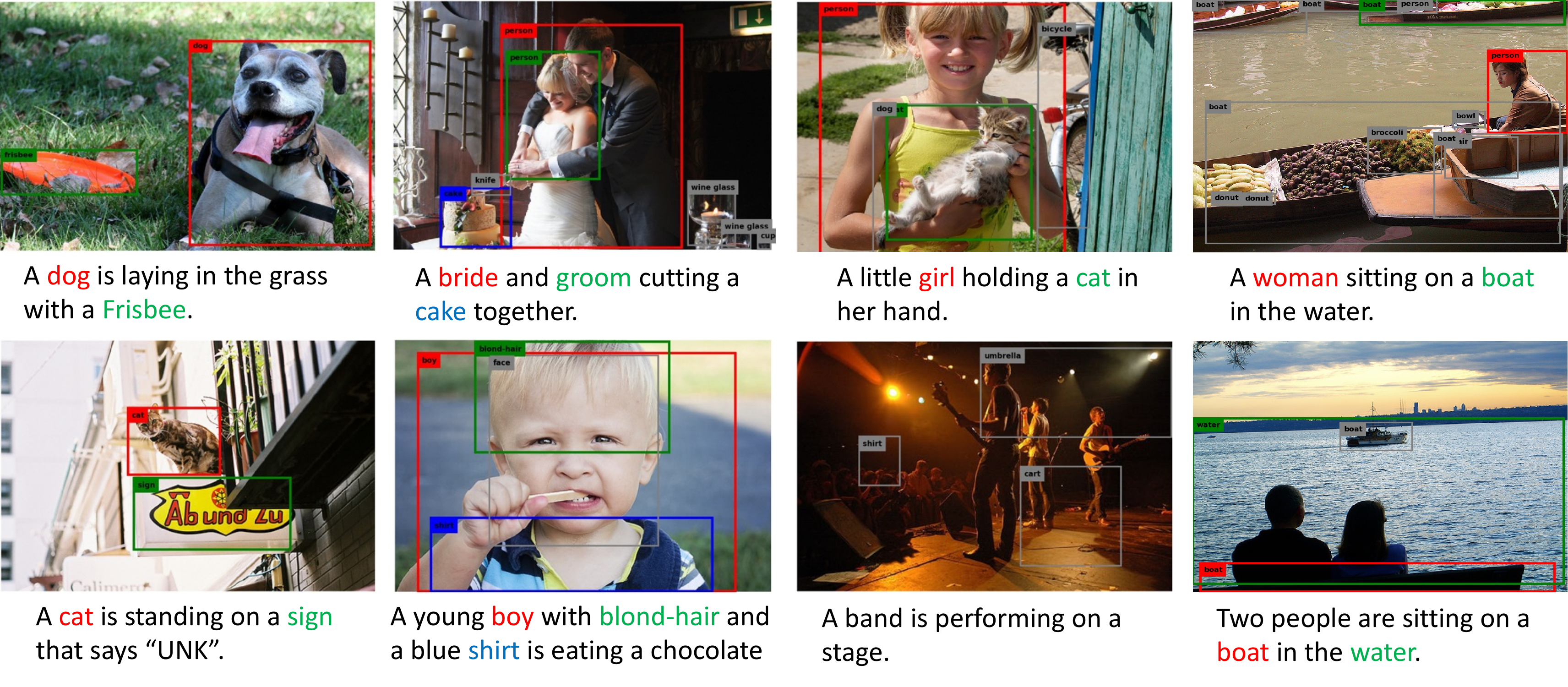}
 \caption{Generated captions and corresponding visual grounding regions on the standard image captioning task (Top: COCO, Bottom: Flickr30k). Different colors show a correspondence between the visual words and grounding regions. Grey regions are the proposals not selected in the caption. First 3 columns show success and last column shows failure cases (words are grounded in the wrong region).}
\label{fig:standard}
\vspace{-2mm}
\end{figure*}

\textbf{Training details.} In our experiments, we use a two layer LSTM with hidden size $1024$. The number of hidden units in the attention layer and the size of the input word embedding are $512$. We use the Adam \cite{kingma2014adam} optimizer with an initial learning rate of $5 \times 10^{-4}$ and anneal the learning rate
by a factor of $0.8$ every three epochs. We train the model up to 50 epochs with early stopping. Note that we do not finetune the CNN network during training. We set the batch size to be 100 for COCO \cite{lin2014microsoft} and 50 for Flickr30k \cite{plummer2015flickr30k}. 

%% file: experiment.tex
\section{Experimental Results}
\label{sec:experiment}
\textbf{Datasets.} We experiment with two datasets. Flickr30k Entities \cite{plummer2015flickr30k} contains 275,755 bounding boxes from 31,783
images associated with natural language phrases. Each image is annotated with 
5 crowdsourced captions. For each annotated phrase in the caption, we identify visual words by selecting the inner most NP (noun phrase) tag from the Stanford part-of-speech tagger~\cite{chen2014fast}.
We use Stanford Lemmatization Toolbox \cite{manning-EtAl:2014:P14-5} to
get the base form of the entity words resulting in 2,567 unique words.

COCO \cite{lin2014microsoft} contains 82,783, 40,504 and 40,775 images for training, validation and testing 
respectively. Each image has around 5 crowdsourced captions. Unlike Flickr30k Entities, COCO does not have bounding box annotations associated with specific phrases or entities in the caption. To identify visual words, we manually constructed an object category to word mapping that maps object categories like $<$person$>$ to a list of potential fine-grained labels like [``child'', ``baker'', ...]. This results in 80 categories with a total of 413 fine-grained classes. See supp. for details. 

\begin{table}[t]\footnotesize
\setlength\tabcolsep{3.2 pt}
\center
  \begin{tabular}{l c c c c c}
    \toprule
    Method &  BLEU1 & BLEU4 & METEOR & CIDEr & SPICE\\
    \midrule
    Hard-Attention \cite{Xu2015show} & 66.9 & 19.9 & 18.5 & - & - \\
    ATT-FCN  \cite{you2016image} & 64.7 & 23.0 & 18.9 & - & - \\
    Adaptive  \cite{lu2016knowing} & 67.7 & 25.1 & 20.4 & 53.1 & 14.5\\
    \midrule
    NBT & \textbf{69.0} & \textbf{27.1} & \textbf{21.7} & \textbf{57.5} & \textbf{15.6} \\ 
    NBT$^{\mathrm{oracle}}$ & 72.0 & 28.5 & 23.1 & 64.8 & 19.6 \\
\bottomrule
 \end{tabular}
 \vspace{1mm}
\small{\caption{Performance on the test portion of Karpathy~\etal~\cite{karpathy2015deep}'s splits on Flickr30k Entities dataset. 
\label{tab:flickr30k-standard}
}}
\vspace{-3mm}
\end{table}

\textbf{Detector pre-training.} 
We use open an source implementation \cite{jjfaster2rcnn} of Faster-RCNN \cite{ren2015faster} to train the detector. For Flickr30K Entities, we use visual words that occur at least 100 times as detection labels, resulting in a total of $460$ detection labels. Since detection labels and visual words have a one-to-one mapping, we do not have fine-grained classes for the Flickr30K Entities dataset -- the caption refinement process only determines the plurality of detection labels. For COCO, ground truth detection annotations are used to train the object detector. 

\textbf{Caption pre-processing.} We truncate captions longer than 16 words for both COCO and Flickr30k Entities dataset. We then build a vocabulary of words that occur at least 5 times in the training set, resulting in 9,587 and 6,864 words for COCO and Flickr30k Entities, respectively. 

\begin{table}[t]\footnotesize
\setlength\tabcolsep{3.5 pt}
\center
  \begin{tabular}{l c c c c c}
    \toprule
    Method & BLEU1 & BLEU4 & METEOR & CIDEr & SPICE \\
    \midrule    
    Adaptive \cite{lu2016knowing} & 74.2 & 32.5 & 26.6 & \textbf{108.5} & 19.5 \\
    Att2in \cite{rennie2016self} & - & 31.3 & 26.0 & 101.3 & - \\
   Up-Down \cite{Anderson2017up-down} & 74.5 & 33.4 & 26.1 & 105.4 & 19.2 \\
   \midrule
    Att2in$^*$ \cite{rennie2016self} & - & 33.3 & 26.3 & 111.4 & - \\
   Up-Down$^\dagger$ \cite{Anderson2017up-down} & 79.8 & 36.3 & 27.7 & 120.1 & 21.4\\
    \midrule
       NBT  & \textbf{75.5} & \textbf{34.7} & \textbf{27.1} & 107.2 & \textbf{20.1} \\
    NBT$^{\mathrm{oracle}}$ & 75.9 & 34.9 & 27.4 & 108.9 & 20.4 \\
\bottomrule
 \end{tabular}
 \vspace{1mm} 
\small{\caption{Performance on the test portion of Karpathy~\etal~\cite{karpathy2015deep}'s splits on COCO dataset. ${*}$ directly optimizes the CIDEr Metric, $\dagger$ uses better image features, and are thus not directly comparable.
\label{tab:coco-standard}
}}
\vspace{-3mm}
\end{table}

\subsection{Standard Image Captioning}

For standard image captioning, we use splits from Karpathy~\etal~\cite{karpathy2015deep} on COCO/Flickr30k. We report results using the COCO captioning evaluation toolkit \cite{lin2014microsoft}, which reports the widely used automatic evaluation metrics, BLEU \cite{papineni2002bleu}, METEOR \cite{denkowski2014meteor}, CIDEr \cite{vedantam2015cider} and SPICE \cite{anderson2016spice}. 

We present our methods trained on different object detectors: Flickr and COCO. 
We compare our approach (referred to as NBT) to recently proposed Hard-Attention \cite{Xu2015show}, ATT-FCN \cite{you2016image} and Adaptive \cite{lu2016knowing} on Flickr30k, and Att2in \cite{rennie2016self}, Up-Down~\cite{Anderson2017up-down} on COCO. 
Since object detectors have not yet achieved near-perfect accuracies on these datasets, we also report the performance of our model under an oracle setting, where the ground truth object region and category is also provided during test time. (referred to as NBT$^\mathrm{oracle}$)
This can be viewed as the upper bound of our method when we have perfect object detectors.

\begin{figure*}[t]
 \centering 
 \includegraphics[width=1\linewidth, scale=1, trim={0in 0in 0.0in 0.0in}, clip]{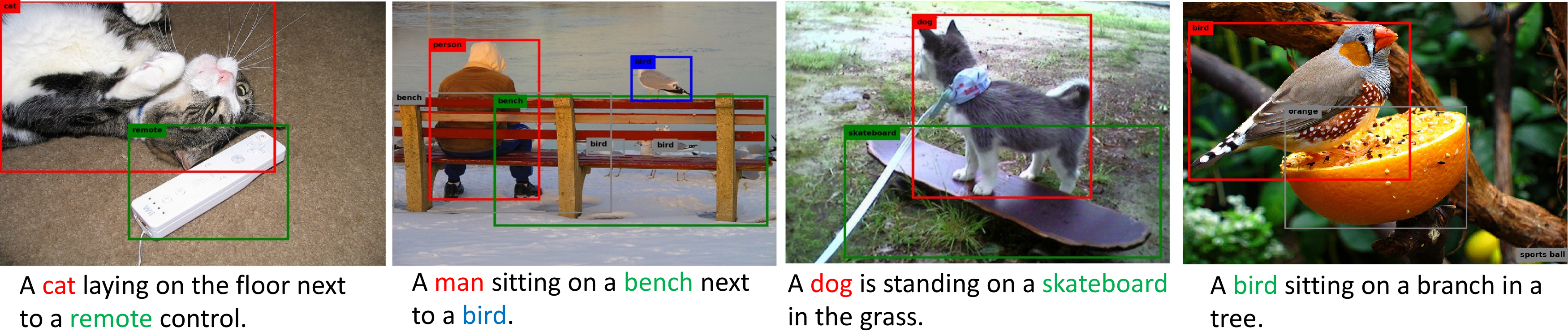}
 \caption{Generated captions and corresponding visual grounding regions for the robust image captioning task. ``cat-remote'', ``man-bird'', ``dog-skateboard'' and ``orange-bird'' are co-occurring categories excluded in the training split. First 3 columns show success and last column shows failure case (orange was not mentioned).}
\label{fig:robust}
\vspace{-4mm}
\end{figure*}

Table~\ref{tab:flickr30k-standard} shows results on the Flickr30k dataset. We see that our method achieves state of the art on all automatic evaluation metrics, outperforming the previous state-of-art model Adaptive \cite{lu2016knowing} by 2.0 and 4.4 on BLEU4 and CIDEr. 
When using ground truth proposals, NBT$^\mathrm{oracle}$ significantly outperforms previous methods, improving 5.1 on SPICE, which implies that our method could further benefit from improved object detectors. 

Table~\ref{tab:coco-standard} shows results on the COCO dataset. Our method outperforms 4 out of 5 automatic evaluation metrics compared to the state of the art~\cite{rennie2016self,lu2016knowing,Anderson2017up-down} without using better visual features or directly optimizing the CIDEr metric. Interestingly, the NBT$^\mathrm{oracle}$ has little improvement over NBT. We suspect the reason is that explicit ground truth annotation is absent for visual words. Our model can be further improved with explicit co-reference supervision where the ground truth location annotation of the visual word is provided. Fig.~\ref{fig:standard} shows qualitative results on both datasets. We see that our model learns to correctly identify the visual word, and ground it in image regions even under weak supervision (COCO). Our model is also robust to erroneous detections and produces correct captions (3rd column). 

\begin{table}[t]\footnotesize
\setlength\tabcolsep{3.5 pt}
\center
  \begin{tabular}{l c c c c c}
    \toprule
    Method & BLEU4 & METEOR & CIDEr & SPICE & Accuracy\\
    \midrule
    Att2in \cite{rennie2016self} & 31.5 & 24.6 & 90.6 & 17.7 & 39.0  \\
    Up-Down \cite{Anderson2017up-down} & 31.6 & 25.0 & 92.0 & 18.1 & 39.7 \\ 
    \midrule
       NBT  & \textbf{31.7} & \textbf{25.2} & \textbf{94.1} & \textbf{18.3} & \textbf{42.4} \\
    NBT$^{\mathrm{oracle}}$  & 31.9 & 25.5 & 95.5 & 18.7 & 45.7 \\
\bottomrule
 \end{tabular}
 \vspace{1mm}
\small{\caption{Performance on the test portion of the robust image captioning split on COCO dataset. 
\label{tab:coco-robust}
}}
\vspace{-3mm}
\end{table}

\begin{figure*}[t]
 \centering 
 \includegraphics[width=1\linewidth, scale=1, trim={0in 0in 0.0in 0.0in}, clip]{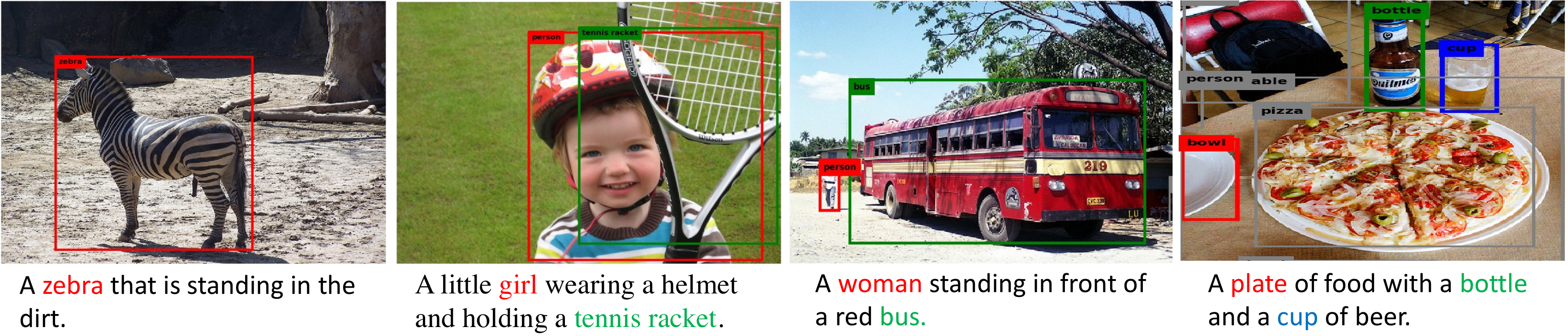}
 \caption{Generated captions and corresponding visual grounding regions for the novel object captioning task. ``zebra'', ``tennis racket'', ``bus'' and ``pizza'' are categories excluded in the training split. First 3 columns show success and last column shows a failure case.}
\label{fig:novel}
\vspace{-1mm}
\end{figure*}

\begin{table*}[t]\footnotesize
\setlength\tabcolsep{3.5pt}
\center
  \begin{tabular}{l c c c c c c c c c c c c c c c}
    \toprule
    \multicolumn{1}{c}{} & \multicolumn{12}{c}{Out-of-Domain Test Data}  & \multicolumn{3}{c}{In-Domain Test Data}\\
   \cmidrule(r){2-13}
   \cmidrule(r){14-16}    
    Method & bottle & bus & couch & microwave & pizza & racket & suitcase & zebra & Avg & SPICE & METEOR & CIDEr & SPICE & METEOR & CIDER\\
    \midrule
    DCC \cite{anne2016deep} & 4.6 & 29.8 & 45.9 & 28.1 & 64.6 & 52.2 & 13.2 & 79.9 & 39.8 & 13.4 & 21.0 & 59.1 & 15.9 & 23.0 & 77.2\\
    NOC \cite{venugopalan2016captioning} & 17.8 & 68.8 & 25.6 & 24.7 & 69.3 & 68.1 & 39.9 & 89.0 & 49.1 & - & 21.4 & - & - & - & - \\
    C-LSTM \cite{yao2017incorporating} & 29.7 & 74.4 & 38.8 & 27.8 & 68.2 & 70.3 & 44.8 & 91.4 & 55.7 & - & 23.0 & - & - & - & - \\
    Base+T4 \cite{anderson2016guided} & 16.3 & 67.8 & 48.2 & 29.7 & 77.2 & 57.1 & 49.9 & 85.7 & 54.0 & 15.9 & 23.3 & 77.9 &  18.0 & 24.5 & 86.3\\
    \midrule
    NBT$^*$+G & 7.1 & 73.7 & 34.4 & 61.9 & 59.9 & 20.2 & 42.3 & 88.5 & 48.5 & 15.7 & 22.8 & 77.0 & 17.5 & 24.3 & 87.4 \\
    NBT$^\dagger$+G & 14.0 & 74.8 & 42.8 & 63.7 & 74.4 & 19.0 & 44.5 & 92.0 & 53.2 & 16.6 & 23.9 & 84.0 & \textbf{18.4} & 25.3 & 94.0 \\
    NBT$^\dagger$+T1  & 36.2 & 77.7 & 43.9 & 65.8 & 70.3 &  19.8 & 51.2 & 93.7 & 57.3 & 16.7 & 23.9 & 85.7 & \textbf{18.4} & \textbf{25.5} & \textbf{95.2} \\
    NBT$^\dagger$+T2  & \textbf{38.3} & \textbf{80.0} & \textbf{54.0} & \textbf{70.3} & \textbf{81.1} & \textbf{74.8} & \textbf{67.8} & \textbf{96.6} & \textbf{70.3} & \textbf{17.4} & \textbf{24.1} & \textbf{86.0} & 18.0 & 25.0 & 92.1 \\
\bottomrule
 \end{tabular}
 \vspace{1mm}
\small{\caption{Evaluation of captions generated using the proposed method. G means greedy decoding, and T1$-$2 means using constrained beam search \cite{anderson2016guided} with 1$-$2 top detected concepts. $*$ is the result using VGG-16 \cite{simonyan2014very} and $\dagger$ is the result using ResNet-101.}}
\label{tab:coco-noc-f1}
\vspace{-3mm}
\end{table*} 

\subsection{Robust Image Captioning}

To quantitatively evaluate image captioning models for novel scene compositions, we present a new split of the COCO dataset, called the robust-COCO split. This new split is created by re-organizing the train and val splits of the COCO dataset such that the distribution of co-occurring objects in train is different from test. We also present a new metric to evaluate grounding.  

\textbf{Robust split.} To create the new split, we first identify entity words that belong to the 80 COCO object categories by following the same pre-processing procedure. 
For each image, we get a list of object categories that are mentioned in the caption. We then calculate the co-occurrence statistics for these 80 object categories. 
Starting from the least co-occurring category pairs, we greedily add them to the test set and ensure that for each category, at least half the instances of each category are in the train set. As a result, there are sufficient examples from each category in train, but at test time we see novel compositions (pairs) of categories. Remaining images are assigned to the training set.
The final split has 110,234/3,915/9,138 images in train/val/test respectively.

\textbf{Evaluation metric.} 
To evaluate visual grounding on the robust-COCO split, we want a metric that indicates whether or not a generated caption includes the new object combination. Common automatic evaluation metrics such as BLEU \cite{papineni2002bleu} and CIDEr \cite{vedantam2015cider} measure the overall sentence fluency. We also measure whether the generated caption contains the novel co-occurring categories that exist in the ground truth caption. A generated caption is deemed 100\% accurate if it contains at least one mention of the \textit{compositionally novel} category-pairs in any ground truth annotation that describe the image.

\textbf{Results and analysis.} 
We compare our method with state of the art Att2in~\cite{rennie2016self} and Up-Down \cite{Anderson2017up-down}. These are implemented using the open source implementation from \cite{Luo2017} that can replicate results on Karpathy's split. We follow the experimental setting from \cite{rennie2016self} and train the model using the robust-COCO train set. Table~\ref{tab:coco-robust} shows the results on the robust-COCO split. As we can see, all models perform worse on the robust-COCO split than the Karpathy's split by 2$\sim$3 points in general. Our method outperforms the previous state of the art methods on all metrics, outperforming Up-Down \cite{Anderson2017up-down} by 2.7 on the proposed metric. The oracle setting (NBT$^\mathrm{oracle}$) has consistent improvements on all metrics, improving 3.3 on the proposed metric. 

Fig.~\ref{fig:robust} shows qualitative results on the robust image captioning task. Our model successfully produces a caption with novel compositions, such as ``cat-remote'', ``man-bird'' and ``dog-skateboard'' to describe the image. The last column shows failure cases where our model didn't select ``orange'' in the caption. We can force our model to produce a caption containing ``orange'' and ``bird'' using constrained beam search \cite{anderson2016guided}, further illustrated in Sec.~\ref{sec:noc_exp}.

\subsection{Novel Object Captioning}
\label{sec:noc_exp}
Since our model directly fills the ``slotted'' caption template with the  concept, it can seamlessly generate descriptions for out-of-domain images. 
We replicated an existing experimental design \cite{anne2016deep} on COCO which excludes all the image-sentence pairs that contain at least one of eight objects in COCO. The excluded objects are `bottle', ``bus'', ``couch'',  ``microwave'', ``pizza'', ``racket'', ``suitcase'' and ``zebra''. We follow the same splits for training, validation, and testing as in prior work \cite{anne2016deep}. We use Faster R-CNN in conjunction with ResNet-101 which is pre-trained on COCO train split as the detection model. Note that we do not pre-train the language model using COCO captions as in \cite{anne2016deep,venugopalan2016captioning,yao2017incorporating}, and simply replace the novel object's word embedding with an existing one which belongs to the same super-category in COCO (e.g., bus $\leftarrow$ car). 

Following \cite{anderson2016guided}, the test set is split into in-domain and out-of-domain subsets. We report F1 as in \cite{anne2016deep}, which checks if the specific excluded object is mentioned in the generated caption. To evaluate the quality of the generated caption, we use SPICE, METEOR and CIDEr metrics and the scores on out-of-domain test data are macro-averaged across eight excluded categories. For consistency with previous work \cite{Anderson2017up-down}, the inverse document frequency statistics used by CIDEr are determined across the entire test set. 

As illustrated in Table~\ref{tab:coco-noc-f1}, simply using greedy decoding, our model (NBT$^*$+G) can successfully caption novel concepts with minimum changes to the model. When using ResNet-101 and constrained beam search \cite{anderson2016guided}, our model significantly outperforms prior works under F1 scores, SPICE, METEOR, and CIDEr, across both out-of-domain and in-domain test data. Specifically, NBT$^\dagger$+T2 outperforms the previous state-of-art model C-LSTM by 14.6\% on average F1 scores. From the category F1 scores, we can see that our model is less likely to select small objects, e.g. ``bottle'', ``racket'' when only using the greedy decoding. Since the visual words are grounded at the object-level, by using \cite{anderson2016guided}, our model was able to significantly boost the captioning performance on out-of-domain images. Fig.~\ref{fig:novel} shows qualitative novel object captioning results. Also see rightmost example in Fig.~\ref{fig:demo1}.

%% file: conclusion.tex
\section{Conclusion}

In this paper, we introduce Neural Baby Talk, a novel image captioning framework that produces natural language explicitly grounded in entities object detectors find in images. Our approach is a two-stage approach that first generates a hybrid template that contains a mix of words from a text vocabulary as well as slots corresponding to image regions. It then fills the slots based on categories recognized by object detectors in the image regions. We also introduce a robust image captioning split by re-organizing the train and val splits of the COCO dataset. Experimental results on standard, robust, and novel object image captioning tasks validate the effectiveness of our proposed approach. 

\small{\textbf{Acknowledgements} }
\small{This work was funded in part by: NSF CAREER awards to DB, DP; ONR YIP awards to DP, DB; ONR Grants N00014-14-1-\{0679,2713\}; PGA Family Foundation award to DP; Google FRAs to DP, DB; and Amazon ARAs to DP, DB; DARPA XAI grant to DB, DP. 
}

%% file: supp.tex
\section{Appendix: COCO Fine-grained Categories}
The COCO \cite{lin2014microsoft} dataset does not have bounding box annotations associated with specific phrases or entities in the caption. We use category level detection annotations and create a category mapping list that maps the object categories like $<$Person$>$ to a list of potential fine-grained labels like [``child'', ``man'', ``baker'',...]. We first use the Stanford lemmatization toolbox \cite{manning-EtAl:2014:P14-5} to get the base form of the entity words in the caption. For each category class, we retrieve the top 200 similar words in the WordVec \cite{mikolov2013efficient} space. We then manually verify each word in the list, resulting in 413 fine-grained classes. A complete list of the fine-grained class for each object category can be found in Table~\ref{tab:class_list} and Table~\ref{tab:class_list2}.

\begin{table*}[b]\footnotesize
\center
  \begin{tabular}{l l}
    \toprule
\textbf{Object category} & \textbf{Fine-grained class} \\
\midrule
$<$person$>$ & person, girl, boy, man, woman, kid, child, chef, baker, people, adult, rider, children, baby, worker, passenger, sister, biker, policeman, \\
& officer, lady, cowboy, bride, groom, male, female, guy, traveler, mother, father, gentleman, pitcher, player, skier, snowboarder, \\
& skater, skateboarder, foreigner, caller, offender, coworker, trespasser, patient, politician, soldier, grandchild, serviceman, walker, \\
& drinker, doctor, bicyclist, thief, buyer, teenager, student, camper, driver, solider, hunter, shopper, villager, cop
\\
$<$bicycle$>$ & bicycle, bike, unicycle, minibike, trike \\
$<$car$>$ & car, automobile, van, minivan, sedan, suv, hatchback, cab, jeep, coupe, taxicab, limo, taxi \\
$<$motorcycle$>$ &  motorcycle, scooter,  motor bike, motor cycle, motorbike, moped \\
$<$airplane$>$ & airplane, jetliner, plane, air plane, monoplane, aircraft, jet, airbus, biplane, seaplane
bus, minibus, trolley \\
$<$bus$>$ &bus, minibus, schoolbus, trolley \\
$<$train$>$ &train, locomotive, tramway, caboose \\
$<$truck$>$ & truck, pickup, lorry, hauler, firetruck\\
$<$boat$>$ &boat, ship, liner, sailboat, motorboat, dinghy, powerboat, speedboat, canoe, skiff, yacht, kayak, catamaran, pontoon, houseboat, vessel, \\
& rowboat, trawler, ferryboat, watercraft, tugboat, schooner, barge, ferry, sailboard, paddleboat, lifeboat, freighter, steamboat, riverboat, \\
& surfboard, battleship, steamship \\
$<$traffic light$>$ & traffic light, street light, traffic signal, stop light, streetlight, stoplight\\
$<$fire hydrant$>$ & fire hydrant, hydrant\\
$<$stop sign$>$ &stop sign, street sign\\
$<$parking meter$>$ &parking meter\\
$<$bench$>$ &bench, pew\\
$<$cat$>$ &cat, kitten, feline, tabby\\
$<$dog$>$ &dog, puppy, beagle, pup, chihuahua, schnauzer, dachshund, rottweiler, canine, pitbull, collie, pug, terrier, poodle, labrador, doggie, \\
& doberman, mutt, doggy, spaniel, bulldog, sheepdog, weimaraner, corgi, cocker, greyhound, retriever, brindle, hound, whippet, husky\\
$<$horse$>$ &horse, colt, pony, racehorse, stallion, equine, mare, foal, palomino, mustang, clydesdale, bronc, bronco\\
$<$sheep$>$ &sheep, lamb, goat, ram, cattle, ewe\\
$<$cow$>$ &cow, cattle, oxen, ox, calf, ewe, holstein, heifer, buffalo, bull, zebu, bison\\
$<$elephant$>$ &elephant\\
$<$bear$>$ &bear, panda\\
$<$zebra$>$ &zebra\\
$<$giraffe$>$ &giraffe\\
$<$backpack$>$ &backpack, knapsack\\
$<$umbrella$>$ &umbrella\\
$<$handbag$>$ &handbag, handbag, wallet, purse, briefcase\\
$<$tie$>$ &tie\\
$<$suitcase$>$ &suitcase, suit case, luggage\\
$<$frisbee$>$ &frisbee\\
$<$skis$>$ &skis, ski\\
$<$snowboard$>$ &snowboard\\
$<$sports ball$>$ & sports ball, baseball, ball, football, soccer, basketball, softball, volleyball, pinball, fastball, racquetball \\
$<$kite$>$ &kite\\
$<$baseball bat$>$ &baseball bat\\
$<$baseball glove$>$ &baseball glove\\
$<$skateboard$>$ &skateboard\\
$<$surfboard$>$ & surfboard, longboard, skimboard, shortboard, wakeboard\\
$<$tennis racket$>$ &tennis racket\\
$<$bottle$>$ &bottle\\
\bottomrule
 \end{tabular}
 \vspace{1pt}
\small{\caption{COCO category mapping list for visual words. 
\label{tab:class_list}
}}
\end{table*}

\begin{table*}[t]\footnotesize
\center
  \begin{tabular}{l l}
    \toprule
\textbf{Object category} & \textbf{Fine-grained class} \\
\midrule
$<$wine glass$>$ &wine glass\\
$<$cup$>$ &cup\\
$<$fork$>$ &fork\\
$<$knife$>$ &knife, pocketknife, knive\\
$<$spoon$>$ &spoon\\
$<$bowl$>$ &bowl, container, plate\\
$<$banana$>$ &banana\\
$<$apple$>$ &apple\\
$<$sandwich$>$ & sandwich, burger, sub, cheeseburger, hamburger \\
$<$orange$>$ &orange, lemons\\
$<$broccoli$>$ &broccoli\\
$<$carrot$>$ &carrot\\
$<$hot dog$>$ &hot dog\\
$<$pizza$>$ &pizza\\
$<$donut$>$ &donut, doughnut, bagel\\
$<$cake$>$ &cake,  cheesecake, cupcake, shortcake, coffeecake, pancake \\
$<$bird$>$ & bird, ostrich, owl, seagull, goose, duck, parakeet, falcon, robin, pelican, waterfowl, heron, hummingbird, mallard, finch, pigeon, sparrow, \\
& seabird, osprey, blackbird, fowl, shorebird, woodpecker, egret, chickadee, quail, bluebird, kingfisher, buzzard, willet, gull, swan, bluejay, \\
& flamingo, cormorant, parrot, loon, gosling, waterbird, pheasant, rooster, sandpiper, crow, raven, turkey, oriole, cowbird, warbler, magpie, \\
& peacock, cockatiel, lorikeet, puffin, vulture, condor, macaw, peafowl, cockatoo, songbird \\
$<$chair$>$ &chair, seat, recliner, stool \\
$<$couch$>$ &couch, sofa, recliner, futon, loveseat, settee, chesterfield \\
$<$potted plant$>$ &potted plant, houseplant\\
$<$bed$>$ &bed \\
$<$dining table$>$ &dining table, table\\
$<$toilet$>$ &toilet, urinal, commode, lavatory, potty\\
$<$tv$>$ &tv, monitor, televison, television\\
$<$laptop$>$ &laptop, computer, notebook, netbook, lenovo, macbook \\
$<$mouse$>$ &mouse \\
$<$remote$>$ &remote \\
$<$keyboard$>$ &keyboard \\
$<$cell phone$>$ & cell phone, mobile phone, phone, cellphone, cellphone, telephone, phon, smartphone, iPhone \\
$<$sink$>$ &sink\\
$<$refrigerator$>$ &refrigerator, fridge, refrigerator, fridge, freezer, refridgerator, frig
\\
$<$book$>$ &book\\
$<$clock$>$ &clock\\
$<$vase$>$ &vase\\
$<$scissors$>$ &scissors\\
$<$teddy bear$>$ &teddy bear, teddybear\\
$<$hair drier$>$ &hair drier, hairdryer\\
$<$toothbrush$>$ &toothbrush\\
\bottomrule
 \end{tabular}
 \vspace{1pt}
\small{\caption{COCO category mapping list for visual words (continued).
\label{tab:class_list2}
}}
\end{table*}